\documentclass{article}

\usepackage[preprint]{neurips_2021} 

\usepackage[utf8]{inputenc} % allow utf-8 input
\usepackage[T1]{fontenc}    % use 8-bit T1 fonts
\usepackage{hyperref}       % hyperlinks
\usepackage{url}            % simple URL typesetting
\usepackage{booktabs}       % professional-quality tables
\usepackage{amsfonts}       % blackboard math symbols
\usepackage{nicefrac}       % compact symbols for 1/2, etc.
\usepackage{microtype}      % microtypography
\usepackage{xcolor}         % colors

\usepackage{algorithmicx,algorithm}
\usepackage{CJK}
\usepackage{threeparttable}
\usepackage{booktabs}
\usepackage{algpseudocode}
\usepackage{subfigure}
\usepackage{amsmath}
\usepackage{hyperref}

\usepackage{graphicx} %插入图片的宏包

\title{Graph-MLP: Node Classification  \linebreak
without Message Passing in Graph}

\author{%

   Yang Hu* \\
   Megvii Inc.  \\
   \texttt{huyang02@megvii.com} \\
   \And
   Haoxuan You \\
   Columbia University \\
   \texttt{hy2612@columbia.edu} \\
   \And
   Zhecan Wang \\
   Columbia University \\
   \texttt{zw2627@columbia.edu} \\
   \And
   Zhicheng Wang \\
   Megvii Inc.  \\
   \texttt{wangzhicheng@megvii.com} \\
   \And
   Erjin Zhou \\
   Megvii Inc.  \\
   \texttt{zej@megvii.com} \\
   \And
   Yue Gao \\
   Tsinghua University, Beijing, China \\
   \texttt{gaoyue@tsinghua.edu.cn} \\
}

\begin{document}

\maketitle

\begin{abstract}
Graph Neural Network (GNN) has been demonstrated its effectiveness in dealing with non-Euclidean structural data. Both spatial-based and spectral-based GNNs are relying on adjacency matrix to guide message passing among neighbors during feature aggregation. Recent works have mainly focused on powerful message passing modules, however, in this paper, we show that none of the message passing modules is necessary.  Instead, we propose a pure multilayer-perceptron-based framework, \textbf{Graph-MLP} with the supervision signal leveraging graph structure, which is sufficient for learning discriminative node representation. In model-level, Graph-MLP only includes multi-layer perceptrons, activation function, and layer normalization. In the loss level, we design a neighboring contrastive (\textbf{NContrast}) loss to bridge the gap between GNNs and MLPs by utilizing the adjacency information implicitly. This design allows our model to be lighter and more robust when facing large-scale graph data and corrupted adjacency information. Extensive experiments prove that even without adjacency information in testing phase, our framework can still reach comparable and even superior performance against the state-of-the-art models in the graph node classification task.

% Using integrated adjacency matrix in graph convolution is also the biggest challenge facing larger scale graph. In this paper, we reorganized the core components of GNN and designed a new framework xxxxxxx (xxx) for graph learning task, which contain Graph-MLP model and Neighbor Contrastive Loss (NCS). Firstly, the proposed Graph-MLP consists of multi-layer perceptrons and some activation function and batch normalization operation inside, which allows more flexible size input compared with the former fixed size input.

% GNN is popular in many tasks. but usually they all need the propagate (message passing) operation as a pluggable module to utilize the node's adjacency information. Which needs extra computing compared with MLPs. Could we empower MLP to perform as well as GNNs by removing the redundant propagating operation? We design a neighbor contrastive (NC) loss function without inserting a new module in the network. The loss function can bridge the gap between GNN and MLPs by introducing the adjacency information in the loss. 
  
\end{abstract}

\section{Introduction}

Deep learning has provided the powerful capacity for machine learning tasks in recent years. Besides the data in the euclidean space, there is an increasing number of graph structured data in the non-euclidean domains. To cope with the graph-related tasks that generally exist in different fields, such as computer vision \citep{xu2017scene} \citep{yang2018graph}, traffic \citep{zhang2018gaan} \citep{li2017diffusion}, recommender systems \citep{he2020lightgcn}, Graph Neural Networks (GNNs) have drawn more and more research interests.  

Among those GNNs, how to build an effective and efficient message passing module between neighbors has been the main focus in order to boost the performance. As a pioneer, Graph Convolutional Network (GCN) \citep{kipf2016semi} sums up the neighbor features to obtain the new node feature. Graph Attention Network (GAT) \citep{velivckovic2017graph} learns attention values among connected nodes to adaptive pass the message and update node features, and multiple multi-head attention layers are stacked.  

% \textbf{Computation Efficiency: }

% Even with enhanced performance, these  message passing modules unavoidably jeopardize computation efficiency and lack of robustness when connection information is corrupted in inference.
Besides sophisticated GNN structures and enhanced performance, there has been a trend to design simpler and more efficient GNN structures. For instance, SGC \citep{wu2019simplifying} reduces excess complexity through successively removing activation layer and using the pre-computed $r^{th}$ power of adjacency matrix. However, they all share the de facto design that graph structure information is explicitly utilized in message passing modules to aggregate features. Considering the time consumption, a natural question is whether we can remove the redundant message passing modules.

In this paper, we aim to learn the structural node representation without explicit message passing. We propose a novel alternative to GNNs, Graph-MLP, where we implicitly use supervision signals from node connection information to guide a pure MLP-based model for graph node classification. In Graph-MLP,  linear layers are combined with activation function, layer normalization, and dropout layers to compose our model structure. The message passing between neighbors through feature aggregation is completely abandoned. Instead, a novel \textbf{N}eighboring \textbf{Contrast}ive (\textbf{NContrast}) loss is proposed to implicitly incorporate graph structure into feature transformation. More specifically, for each node, its \textit{i}-hop neighbors are treated as positive samples and other unconnected nodes are regarded as negative samples. The distance between the selected node against the positive/negative would drew/pushed closer/farther. 

The advantages of Graph-MLP over previous GNNs exist in two folds. 1) Higher computation efficiency via bypassing message passing in feed-forward propagation: 
Since our model structure does not require explicit message passing as the conventional graph models, this allows us more freedom in choosing simpler base model \textit{i.e.} MLP. A Simple and light base model is often required in lots of applications.
2) Robustness against corrupted edges during inference: In real-world applications, it often happens that some new samples are not connected to any node in the existing graph, such as "a new social network platform user to a recommendation system". And sometimes, there might be some noise in collected connection information.  Then conventional GNNs are incompetent in providing recommendations due to the vacant or noist adjacency information. However, adjacency connection is not needed in the inference phase of Graph-MLP framework. Because our Ncontrast loss allows Graph-MLP to learn a structural-aware nodes' feature transformation. Thus Graph-MLP can still provide consistent recommendations despite the user node's missing connections.

\begin{figure}[]
\centering
\includegraphics[width=0.9\linewidth]{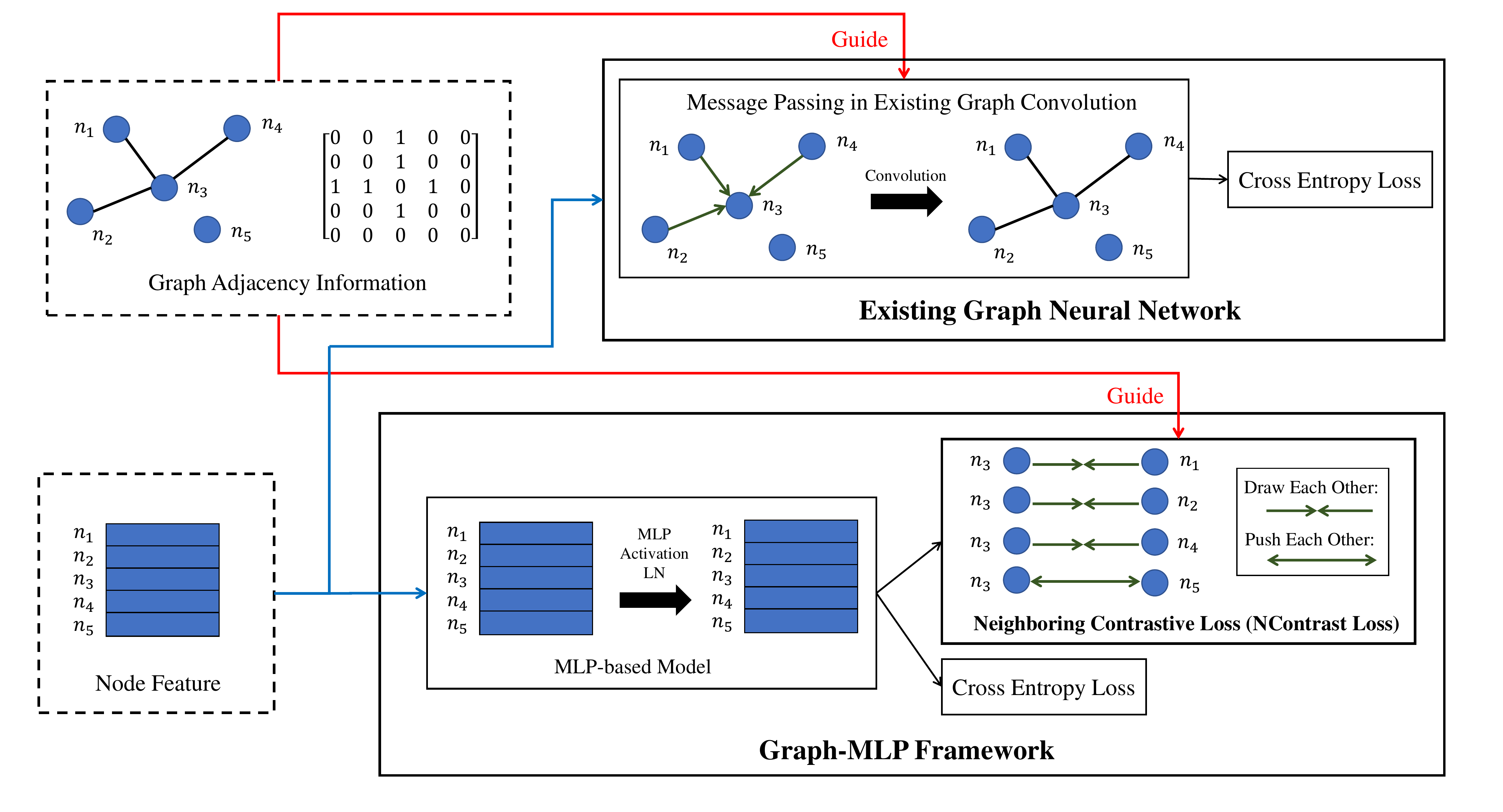}
\caption{Diagram of Graph-MLP. GNNs take the adjacency matrix and node feature as input for message passing on node classification task. Graph-MLP takes node feature as input and utilizes the adjacency matrix in NContrast loss.}
\label{pipeline}
\vspace{-4mm}
\end{figure}

To demonstrate the Graph-MLP's superiority, we conduct extensive experiments on two points: (1) Graph-MLP can achieve comparative performance as GNNs with higher efficiency in three node classification benchmarks. (2) Graph-MLP's robustness in feature transformation during inference with corrupted connection information.

Our contributions are summarized as follows:

\begin{itemize}
\item[$\bullet$] A simple MLP-based graph learning framework (Graph-MLP) without message passing modules. To our best knowledge, it is the first deep learning framework to effectively perform graph node classification task without explicit message passing modules.
\item[$\bullet$] A novel Neighboring Contrastive loss to implicitly incorporate graph structure into node representation learning.
\item[$\bullet$] Experimental results show that Graph-MLP can achieve comparable or even superior performance on node classification tasks with higher efficiency. When adjacency information is corrupted or missing in the inferencing time, Graph-MLP robustly maintains consistent performance.
\end{itemize}

The rest of the paper is structured as follows. Related works are reviewed in Section 2. Necessary preliminaries and Graph-MLP framework are introduced in  Section 3. Extensive experiments are reported in Section 4.

\section{Related Work}

\paragraph{Graph Neural Networks}
GCN \citep{kipf2016semi}  first generalizes  classic  Convolutional  Neural  Net-works  (CNNs)  from  euclidean  domain  to  graph  domain. Then follow-up works can be divided into SpatialGCNs and Spectral GCNs. \citep{you2020design} surveys a wide range of GNN design spaces for different tasks. It also includes pragmatic guidelines in designing GNNs for several tasks. There is also some other form of graph neural networks concentrating on hypergraphs \citep{feng2019hypergraph}.To help GNN learning representations on Riemannian manifolds, \citep{chami2019hyperbolic} and \citep{liu2019hyperbolic} utilize hyperbolic embedding to facilitate learning.

\paragraph{Multilayer Perceptron}
Recently, MLPs are attracting vision reserch to design more general and simple architecture. \citep{tolstikhin2021mlpmixer} \citep{melaskyriazi2021need} design MLP based network and performs as well as prevalent vision networks \citep{dosovitskiy2020image}.
\citep{touvron2021resmlp} proposed residual blocks consisting of feed-forward network and interaction layer.
\citep{ding2021repmlp} use re-parameterization technique to add local prior into MLP to make it powerful for image recognition.
These work pave the way for applying MLP-based model to different research domains. And they demonstrate the potential of MLP-based models against other complicated models.

\paragraph{Contrastive Learning}
Contrastive learning is widely applied in self-supervised learning such as \citep{chen2020simple} \citep{chen2020big} \citep{he2020momentum} \citep{chen2020improved} \citep{grill2020bootstrap}. Contrastive learning is also used in supervised way like \citep{khosla2021supervised}. In natural language processing, \citep{gunel2020supervised} use supervised contrastive learning to help to pre-train a large language model on an auxiliary task. There are also self-supervised learning applied on graph learning. Exhaustive information is referred to \citep{jin2020self}.

\section{Method}

Following GCNs \citep{kipf2016semi}, we introduce our Graph-MLP in the context of node classification task. In this task, a graph consisted of both labeled and unlabled nodes is input into a model, and the output is the prediction of unlabeled nodes.  For a given graph $\mathcal{G}$,  we can represent it as $\mathcal{G}=(\mathcal{V}, \mathbf{A})$, where the vertex set is $\mathcal{V}$  containing nodes $\left\{v_{1}, \ldots, v_{n}\right\}$ and $\mathbf{A} \in \mathbb{R}^{n \times n}$ is the adjacency matrix where $a_{i j}$ denotes the edge weight between nodes $v_{i}$ and $v_{j}$. In the graph $\mathcal{G}$, we denote the node feature matrix as $\mathbf{X} \in \mathbb{R}^{n \times d}$ where the feature vector of each node $v_{i}$ is $\mathrm{x}_{i} \in \mathbb{R}^{d}$. Each node belongs to one category of $C$ classes with a one-hot label, $\mathbf{y}_{i} \in\{0,1\}^{C}$. The node classification task requires the model to give predicted labels of the test nodes.

\subsection{Revisiting GNNs}
In vanilla GCN, MLPs and summation operation over connected nodes are combined together to propagate features among neighbors. 
\begin{equation}
    \mathbf{X}^{(l+1)}=\sigma\left(\widehat{A} \mathbf{X}^{(l)} W^{(l)}\right)
    \label{eq_gcn}
\end{equation}
where $\mathbf{X}^{(0)} = \mathbf{X}$ and $\mathbf{X}^{(l)}$ means the $l^{th}$ layer features. $W^{l}$ is the weight of the single linear layer of the $l^{th}$ layer and $\sigma$ is an activation function. The normalized adjacency matrix $\widehat{A}$ is utilized to pass messages between neighbors.  
\begin{equation}
\widehat{A}=\mathbf{D}^{-\frac{1}{2}} \left(A + I\right) \mathbf{D}^{-\frac{1}{2}}
\end{equation}
where $I$ denotes self-connection, $\mathbf{D}$ is a diagonal matrix and $\mathbf{D}_{(i, i)}=\sum_{j} \widetilde{A}_{(i, j)}$.

In follow-up works, various ways of passing messages between neighbors have been proposed. For instance, GAT employs an attention module to dynamically learn the attention weight of message passing among neighbors. GGNN \citep{li2015gated} incorporate GRU \citep{chung2014empirical} into message passing process to allow the interaction between memory and current feature.

\begin{figure}[]
\centering
\includegraphics[width=1.0\linewidth]{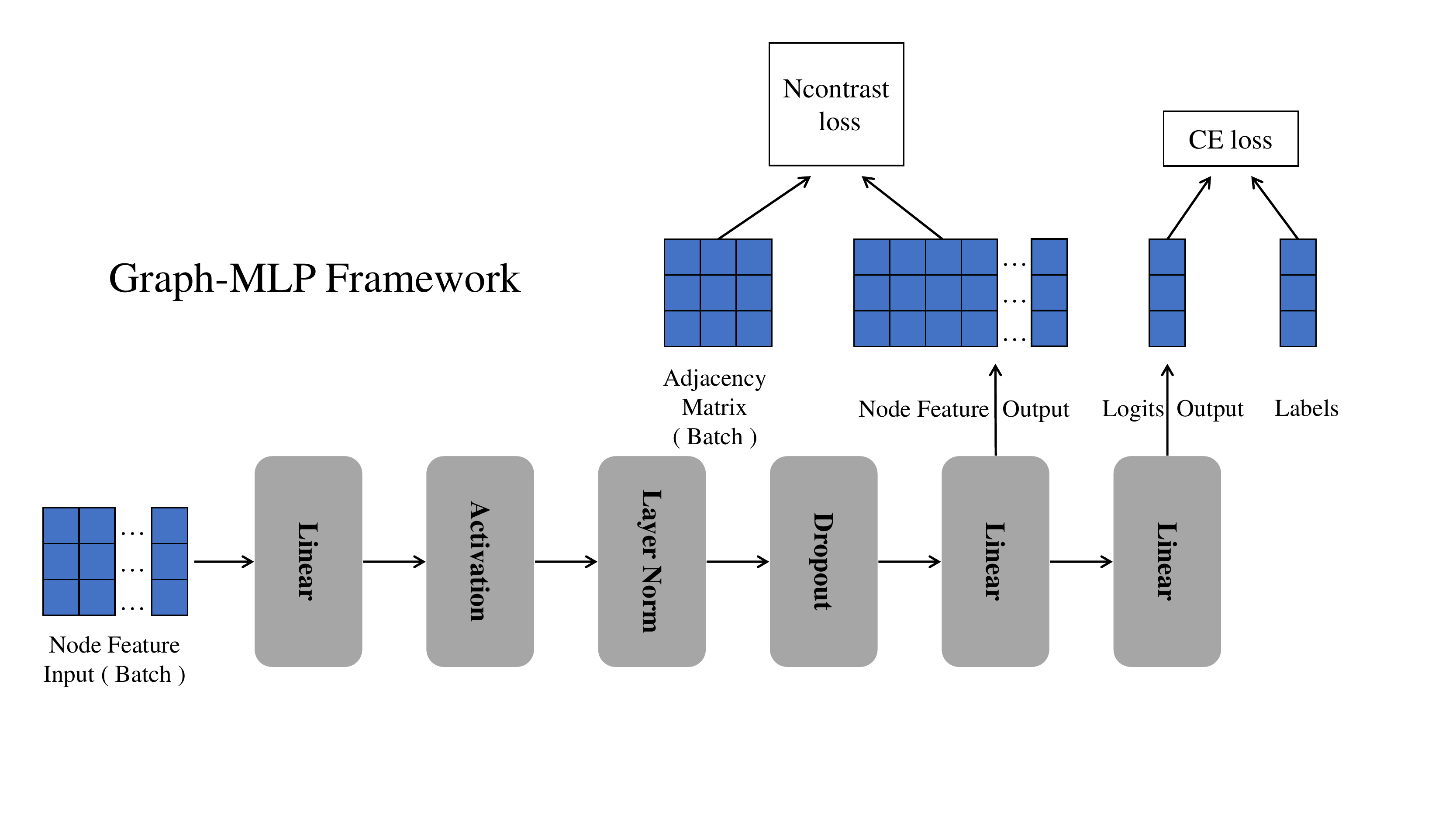}

\caption{The framework of Graph-MLP.}
\label{arch}
\end{figure}

\subsection{Graph-MLP}

Previous graph models heavily focus on leveraging message passing among neighbors. By their design, graph structure is explicitly learned in feed-forward feature propagation. Those sophisticated message passing often lead to complex structure and heavy computation. In this work, we creatively explore a new perspective to learn node feature transformation without explicit message passing. In structure level, despite its simple design, we use a pure MLP-based model. To compensate for its light structure in graph modelling, we combine it with a novel contrastive loss to supervise its learning of node feature transformation.

In the following, we would introduce our proposed framework, Graph-MLP in three perspectives: 1) The MLP-based model structure; 2) The \textbf{N}eighbouring \textbf{Contrast}ive (\textbf{NContrast}) loss for supervising node feature transformation. 3) Training and Inferencing of Graph-MLP.

\paragraph{MLP-based Structure}

Following the common design, our MLP-based structure consists of linear layers followed by activation, normalization, and dropout as in Figure \ref{arch}. More specifically, a linear-activation-layer normalization-dropout block is utilized and two following additional linear layers serve as the prediction heads. For activation, we use Gelu \citep{hendrycks2016gaussian}. Layernorm \citep{ba2016layer} is applied instead of batch normalization for the training stability. Dropout \citep{srivastava2014dropout} is to avoid over-fitting.   The second last linear layer is supervised for feature transformation with our proposed NContrast loss and, differently, the last linear layer is guided for learning node classification. The entire model can be formulated as:

\begin{equation}
     \mathbf{X}^{(1)}=Dropout\left(LN\left(\sigma\left(\mathbf{X} W^{0}\right)\right)\right)
\end{equation}
\begin{equation}
    \mathbf{Z} = \mathbf{X}^{(1)}  W^{1}
\end{equation}
\begin{equation}
    \mathbf{Y} = \mathbf{Z}  W^{2}
\end{equation}

where $\mathbf{Z}$ will be used for NContrast loss, and $\mathbf{Y}$ will be used for classification loss. Compared with Equation \ref{eq_gcn} and other GNNs, Graph-MLP does not require the normalized adjacency matrix $\widehat{A}$ and corresponding message passing step.

\paragraph{Neighbouring Contrastive Loss}
For enabling the feature extraction with graph connection information, it is intuitive that connected nodes should be similar to each other and unconnected ones should be far away in feature space. This aligns well with the idea inside contrastive learning. With such motivation, we propose a Neighbouring Contrastive loss which enables an MLP-based model to learn graph node connections without explicit messaging passing modules.

In NContrast loss, for each node, its $r$-hop neighbors are regarded as the positive samples, while the other nodes are sampled as the negative ones. The loss encourages positive samples to be closer to the target node and pushes negative samples away from the target one in terms of feature distance. In details, NContrast loss for the $i^{th}$ node can be formulated as:

\begin{equation}
\ell_{i}=-\log \frac{  \sum_{j=1}^{B} \mathbf{1}_{[j \neq i]}   \gamma_{ij}  \exp \left(\operatorname{sim}\left(\boldsymbol{z}_{i}, \boldsymbol{z}_{j}\right) / \tau\right)}{\sum_{k=1}^{B} \mathbf{1}_{[k \neq i]} \exp \left(\operatorname{sim}\left(\boldsymbol{z}_{i}, \boldsymbol{z}_{k}\right) / \tau\right)}
\label{contrast}
\end{equation}

% In the above Graph-MLP structure, the last two layers serve as the prediction layers corresponding to two separate supervising losses.  Besides extracting the last layer
% s output for cross entropy loss, we also extract the second last layer, $L-1$'s feature $\mathbf{X^{L-1}}\in \mathbb{R}^{B \times d}$ ($d$ is feature dimension of $L-1$ layer) and the adjacency matrix to compute NC loss. The NC loss for the $i^{th}$ node can be written as:

% \begin{equation}
% \ell_{i}=-\log \frac{  \sum_{j=1}^{B} \mathbf{1}_{[j \neq i]}   \gamma_{ij}  \exp \left(\operatorname{sim}\left(\boldsymbol{z}_{i}, \boldsymbol{z}_{j}\right) / \tau\right)}{\sum_{k=1}^{B} \mathbf{1}_{[k \neq i]} \exp \left(\operatorname{sim}\left(\boldsymbol{z}_{i}, \boldsymbol{z}_{k}\right) / \tau\right)}
% \label{contrast}
% \end{equation}

where $\operatorname{sim}$ denotes the cosine similarity and $\tau$ denotes the temperature parameter. $\gamma_{ij}$ means the strength of connection between node $i$ and $j$. We compute it as the $r^{th}$ power of normalized adjacency matrix $\widehat{A}^{r}$: $\gamma_{ij} = \widehat{A}^{r}_{ij}$. $\gamma^{ij}$ gets non-zero values only if node $j$ is the $r$-hop neighbor of node $i$. 

\begin{equation*}
     \gamma_{ij} \begin{cases}
= 0,& \text{node $j$ is the $r$-hop neighbor of node $i$} \\
\ne 0,& \text{node $j$ is not the $r$-hop neighbor of node $i$ }\\
\end{cases}
\label{eq_nc}
\end{equation*}

Besides the NContrast loss, we also have a traditional cross-entropy (CE) loss for node classification, as in \citep{kipf2016semi}. It's noted that NContrast loss is applied after the second last layer and CE loss is on top of the last layer. In total, we define the final loss of Graph-MLP as a combination of both: 

% (e.g., $r$ = 2, As described in \citep{wu2019simplifying}. The 2 order of adjacency matrix is powerful enough to aggregate the node feature from its neighbors.). 

% When training in batches, we first compute the $\widehat{A}^{r}$ in the whole graph. Then we sample a batch of node feature and the batch of adjacency matrix as described in Figure \ref{pipeline}(c). In our node classification task. We define the total loss for backpropagation:
\begin{equation}
    loss_{NC}=\alpha\frac{1}{B}\sum_{i=1}^{B}\ell_{i}
    \label{loss}
\end{equation}
\begin{equation}
    loss_{final} = loss_{CE} + loss_{NC}
\end{equation}
 
where $loss_{CE}$ is the cross entropy loss. $\alpha$ is the weighting coefficient to balance the those two losses. 

% \paragraph{Training and Inference}

% The total loss is trained in an end-to-end manner. 

\begin{algorithm}[t]
\caption{Graph-MLP training} 
%\begin{flushleft}
{\bf Input:} 
Feature matrix $\mathbf{X}$ and $r^{th}$ power of adjacency matrix $\widehat{A}^{r}$. The train/validation/test index. Hyperparameter $B$, $\alpha$, $\tau$. Trainable model parameters $\theta$. Training iterations $T$.
\\
{\bf Output:}
Optimized model parameters $\theta$.
\label{algo1}

%\end{flushleft}
\begin{algorithmic}[1]

\State 

\State \(t \leftarrow 0\)

\While{\(t<T\)}
    \State Sample the input batch node feature matrix $\mathbf{X}[Id_{B}, :]$ and the input batch adjacency matrix $\widetilde{A}[Id_{B},Id_{B}]$
    \State Compute $loss$ with Eq. \ref{loss}
    
    \State Perform backpropagation with $loss$ and update the model parameter $\theta$
\EndWhile
\end{algorithmic}
\end{algorithm}

% --------------------------------------

% (1) the architecture design with usual modules that can be trained end-to-end without GNN's aggregation (convolution/message passing) operation. 

\paragraph{Training }
The entire model is trained in an end-to-end manner. The feedforward of our model does not require adjacency matrix and we only reference graphical structure when computing the loss during training. This allows us much more flexibility against the conventional graph modelling that our framework can be trained with batches without the full graph information \citep{kipf2016semi}. As shown in Algorithm \ref{algo1}, in each batch, we randomly sample $B$ nodes and take the corresponding adjacency information $\widehat{A} \in \mathbb{R} ^{B\times B}$ and node features $\mathbf{X} \in R^{\mathbb{R} \times d}$. It's noted that, in Equation \ref{eq_nc}, for some node $i$, it may happen that no positive sample is in the batch, due to the randomness of batch sampling. In that case, we will remove the loss for node $i$. And we find our model robust to the ratio of positive and negative samples, without specifically tuned ratio. A better batch sampling method and the fusion of Graph-MLP and works like GraphSAGE \citep{hamilton2017inductive} are still remained for future exploration. 

% The batch sampling method we use is quite simple compared with \citep{hamilton2018inductive}\citep{chen2018fastgcn}

\paragraph{Inference }
During the inferencing, the conventional graph modellings \textit{e.g.} GNNs need both the adjacency matrix and node features as inputs. Differently, our MLP-based method only requires node features as input. Hence, when adjacency information is corrupted or missing, Graph-MLP could still deliver consistently reliable results. In the conventional graph modelling, graph information is embedded in adjacency matrix in input. For those models, the learning of graph node transformation heavily relies on internal message passing which is sensitive to connections in each adjacency matrix input. However, our supervision of graphical structure is applied on  the loss level. Thus our framework is able to learn a graph-structured distribution during node feature transformation without feed-forward message passing. This allows our model to be less sensitivity to specific connections during inference.

\section{Experiments}

% \subsection{Performance on Transductive Node Classification}

\subsection{Datasets and Experiment Setting}
In this section, we evaluate the performance of Graph-MLP against the prevalent GNN models on the three popular citation network datasets in node classification as in Table \ref{data_intro_table}. In semi-supervised node classification, $140/500/1\,000$ nodes are used for training/validation/testing in Cora. 120/500/1, 000 nodes are used for training/validation/testing in Citeseer. 60/500/1, 000 nodes are used for training/validation/testing in Pubmed.

\begin{table}
\begin{minipage}{0.45\linewidth}
  \centering
  
\begin{center}
\begin{tabular}{lccc}
% \hline 
\toprule
\text{Dataset} & \text{Cora} & \text{Citeseer} &\text{Pubmed}\\
\midrule
\text { Node } & 2,708 & 3,327 & 19,717  \\
\text { Edge } &  5,429 & 4,732 & 44,338 \\
\text { Feature } & 1,433 & 3,703 &  500 \\
\text { Class } &7 & 6 & 3 \\
% \text { Training node } &7 & 6 & 3 \\
% \text { Validation node } &7 & 6 & 3 \\
% \text { Testing node } &7 & 6 & 3 \\

\hline
\end{tabular}
\end{center}
\caption{Statistics of three citation network datasets.}
\label{data_intro_table}
\end{minipage}
\begin{minipage}{0.45\linewidth}
\begin{center}
\begin{tabular}{l|c|c|c}
\toprule & Cora & Citeseer & Pubmed \\
\midrule
DeepWalk  & $70.7$ & $51.4$ & $76.8$ \\
AdaLNet  & $80.4$ & $68.7$ & $78.1$ \\
LNet  & $79.5$ & $66.2$ & $78.3$ \\
GCN  & $81.5$ & $70.3$ & $79.0$ \\
GAT   & $83.0$ & $72.5 $ & $79.0 $ \\
DGI  & $\mathbf{82.3}$ & $71.8$ & $76.8$ \\
SGC  & $81.0$ & $71.9$ & $78.9$ \\
\midrule
MLP ($\alpha$=0) & $57.8$ & $54.7$ & $73.3$ \\
Graph-MLP  & $79.5$ & $\mathbf{73.1}$ & $\mathbf{79.7}$ \\
\bottomrule

\end{tabular}
\end{center}
\caption{Test accuracy (\%) on citation network datasets. The scores in the table are obtained by averaging over 10 runs for each experiment.
}
\label{semi_results}
\end{minipage}
\end{table}

\textbf{Graph-MLP Network Structure}
We use the network with the same structure as Figure \ref{arch}. We set the hidden dimension $d$=256 in each linear layer. The dropout rate is 0.6 fixed. We use Gelu \citep{hendrycks2016gaussian} as the activation function.

\textbf{Training Setups} 
The training iterations for each data set is fixed with 400. We use Adam \citep{kingma2014adam} as the optimizer. We also sweep the learning rate in the range of [0.001,0.01,0.05,0.1], the weight decay in the range of  [5e-4,5e-3], the batch size in the range of [2000,3000], the temperature parameter $\tau$ in the range of [0.5,1.0,2.0], the $r$ parameter (for computing $r^{th}$ power of adjacency matrix) in the range of [2,3,4] and the $\alpha$, as in Equation \ref{loss}, in the range of [0,1,10,100] (when $\alpha$ is 0, the Graph-MLP degrade into vanilla MLP). 

\textbf{Evaluating Setups} 
We report the final testing results based on the best performed model on validation results. Each result is run 10 times with random initialization. 

% \textbf{Baselines on citation networks node classification:} 

\textbf{Performance on citation networks node classification dataset:}
We compare our results versus several the state-of-the-art graph learning methods including LNet \citep{liao2019lanczosnet}, AdaLNet \citep{liao2019lanczosnet}, DeepWalk \citep{perozzi2014deepwalk}, DGI \citep{velickovic2019deep}, GCN \citep{kipf2016semi} and SGC \citep{wu2019simplifying}. In Table \ref{semi_results}, Graph-MLP can achieve the state-of-the-art performance on Citeseer and Pubmed among the all. Even on Cora, our results are also comparable with other methods. The reason of a slightly lower performance on Cora might be due to the scale of graph. A smaller graph may not be able to provide enough genearl contrastive supervision compared with larger graphs. Compared with the original vanilla MLP (with $\alpha$=0, Graph-MLP behaves the same as original MLP), Graph-MLP improves by 21.7\%, 18.4\%, and 6.4\% respectively on Cora, Citeseer, and Pubmed. This substantial improvement clearly demonstrates the contribution from our proposed NContrast loss.

\subsection{A Deeper Understanding of Graph-MLP}

\textbf{Efficiency of Graph-MLP versus GNNs}
To compare Graph-MLP and GCN \citep{kipf2016semi} in training and testing behavior, we add message passing modules \textit{e.g.} graph convolution before the first and second linear layer in our MLP-based structure to transform it to a two-layer GCN as in \citep{kipf2016semi}. In this way, we can compare the training and testing efficiency fairly with only message passing layer being a variable. For the common hyperparameters of Graph-MLP and GCN, we use a Adam optimizer with the learning rate as 0.01, the weight decay as 5e-4 and the hidden dimension as 256. For our Graph-MLP's specific setting, we set the $\alpha$ to be 1, the batch size to be 2000, $r$ to be 2 in all three datasets. The comparison is evaluated regarding training converge time and inferencing time. It is noted that batch sampling time is also counted in Graph-MLP's training coverage time.

\begin{figure}[]
\centering
\includegraphics[width=0.8\linewidth]{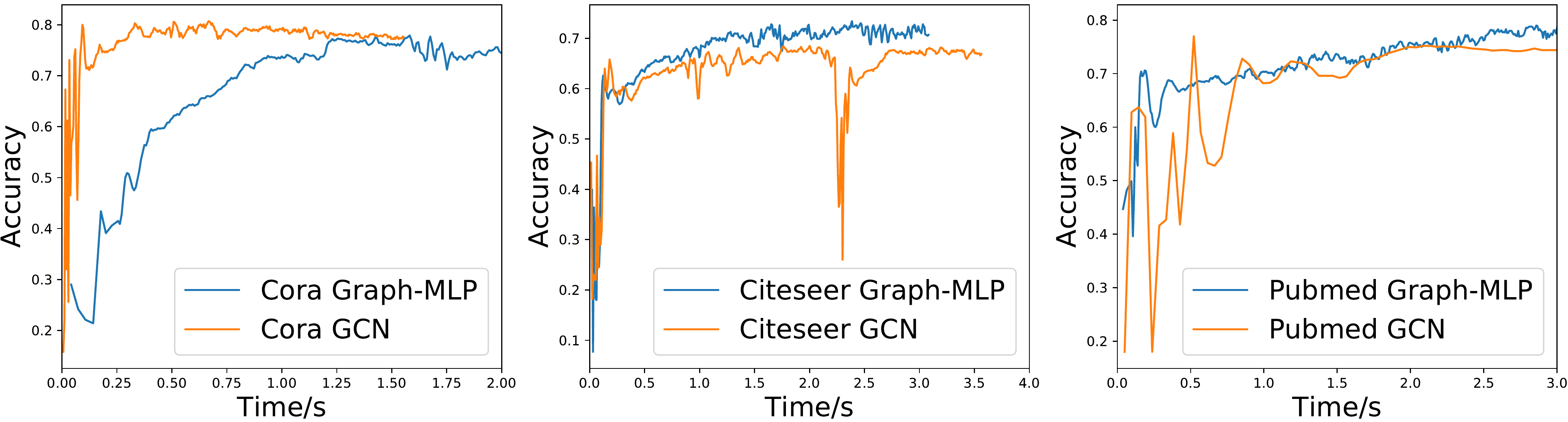}
\caption{Accuracy along training time of Graph-MLP and GCN.}
\label{effi_train}
\end{figure}

 In Figure \ref{effi_train}, we find that Graph-MLP can achieve better results on Citeseer and Pubmed with a quicker converge speed compared with GCN. Additionally, our testing accuracy is more stable and smooth compared with GCN's which demonstrates an inconsistent performance mixed with fluctuations in the testing accuracy, especially as shown in the middle and the rightmost subfigures in Figure. \ref{effi_train}. In Cora, GCN converges earlier, however, the final converged performance of the two is comparable. The reason behind may still be the smaller scale of Cora dataset.  Furthermore, we compare the inferencing time for the testing nodes. GCN needs both the adjacency matrix and the node features to be input, but our Graph-MLP only requires testing node features. We run the inferencing process 100 times and average the recordings to remove the randomness. The time consumption for testing is listed in Table \ref{test_time}. We can see that GCN's inferencing time will increase with the larger number of nodes in the whole graph because of multiplication of the larger adjacency matrix. Nevertheless, with credits to the simiplicity of Graph-MLP, its inferencing time is much faster and only depends on the number of testing nodes. Overall, Graph-MLP shows superior efficiency in both training and inferencing.

\begin{table}[t]
\begin{center}
    
\begin{tabular}{l|c|c|c}
\toprule & Cora & Citeseer & Pubmed \\
\midrule
GCN   & $0.000917$ & $0.001918 $ & $0.010107 $ \\
\midrule
Graph-MLP & $0.000276$ & $0.000262$ & $0.000262$ \\
\bottomrule
\end{tabular}
\end{center}
\caption{Testing time (s) for Graph-MLP and GCN.
}
\label{test_time}
\end{table}

\begin{figure}[]
\centering
\includegraphics[width=1.0\linewidth]{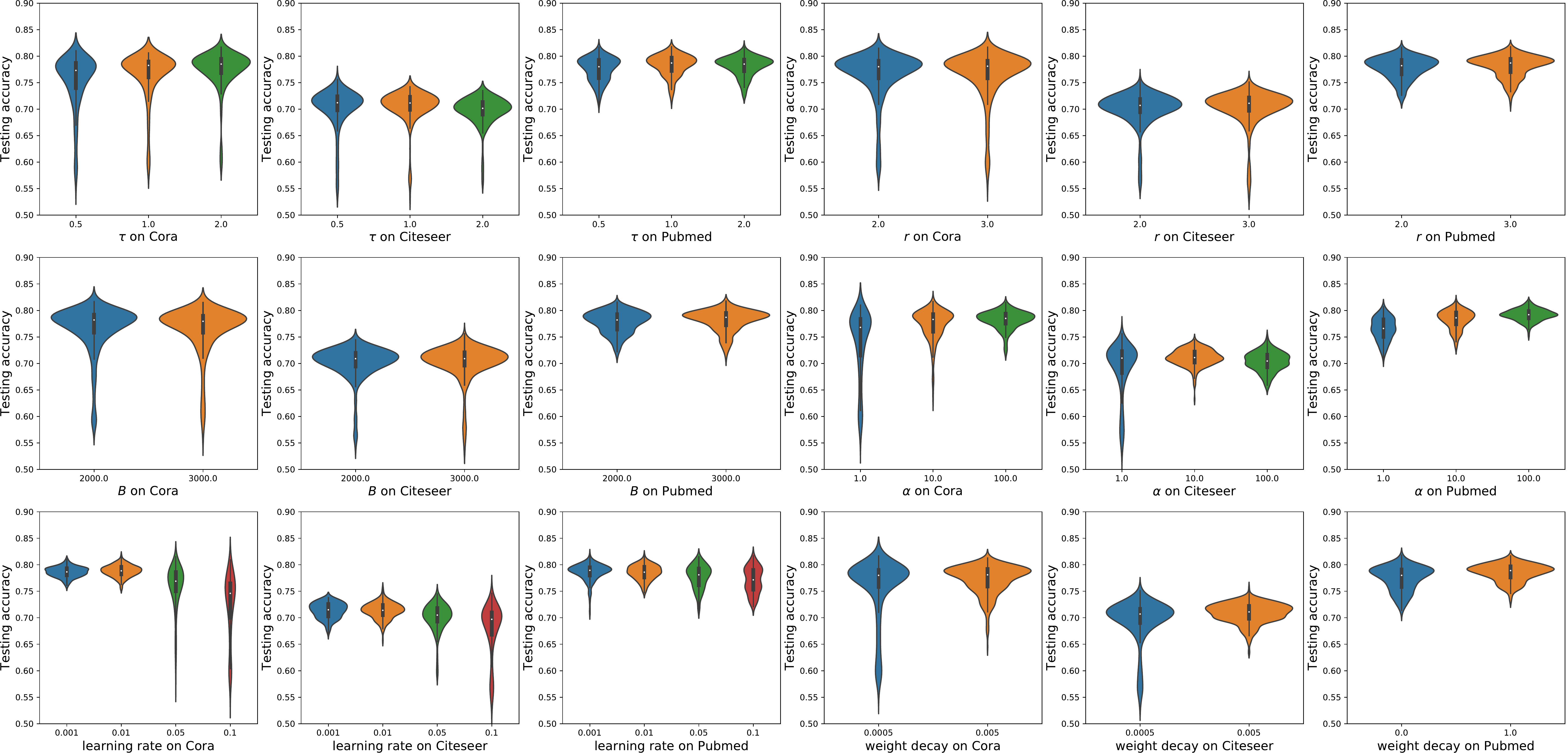}
\caption{Ablation study on the hyperparameters ( including $\tau$, $r$, $B$, $\alpha$, learning rate, and weight decay) on citation network datasets. A hyperparameter with higher and more condensed distribution represents its superiority over its counterpart.}
\label{ablation}
\end{figure}

\textbf{Ablation Study about Hyperparameters}
To have a deeper understanding about the effect of hyperparameters in Graph-MLP, we give an exhaustive analysis on them as in Figure \ref{ablation}. Among the hyperparameters $\tau$, $r$, $B$, $\alpha$, learning rate, and weight decay. We find that: (1) $\tau$, $r$, $B$ are trivial hyperparameters for their values' change doesn't affect the distribution of the results much. In another aspect, it shows that Graph-MLP is robust to the mentioned hyperparameters. (2) As $\alpha$'s value gets bigger from 1 to 10, accuracy gets consistent improvement on Cora and Pubmed. However, with $\alpha$ as 100, performance on Cora and Pubmed drops a little.   On Citeseer, with  $\alpha$ from 1 to 100, the performance keeps improving and gives the best one around 100. (3) Learning rate with 0.01 can usually perform well, which is a recommending default setting. Also, observing that weight decay with 5e-3 is slightly better than 5e-4, we hypothesize that our Graph-MLP may need a stronger regularization.

\begin{figure}[]
\centering
\includegraphics[width=0.9\linewidth]{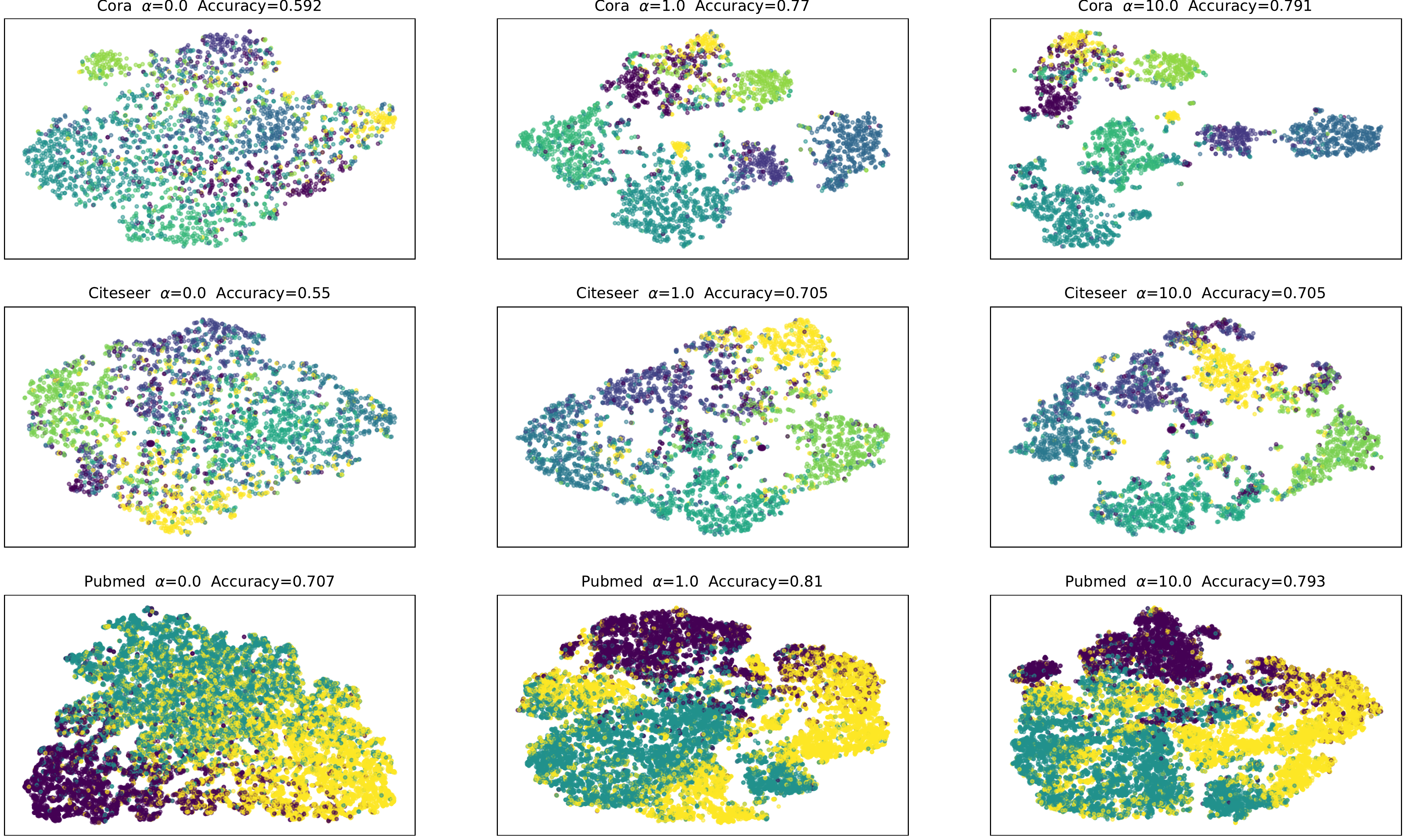}
\caption{t-SNE visualization of feature embedding of citation networks. We plot the whole nodes feature produce from the converged model.}
\label{tsne}
\end{figure}

\textbf{Visualization of Embeddings}
In order to have a visual understanding of how NContrast loss helps vanilla MLP, we visualize the feature embedding of $\mathbf{Z}$ with t-SNE \citep{van2008visualizing}. Here, learning rate is 0.01, weight decay is 5e-4, $\alpha$ ranges from [0.0, 1.0, 10.0], the hidden dimension $d$ is 256, batch size is 2000, $r$ is 2, and $\tau$ is 2. The t-SNE visualization results are plotted as in Figure \ref{tsne}. For example, in Cora, as $\alpha$ grows bigger from 0 to 100, the node embeddings of the same class become cloaser and the embeddings of different classes are pushed farther from each other. Visualized results on Citeseer and Pubmed also reveal the similar effect.

\subsection{Robustness against Corrupted Connection in Inference}
\begin{figure}[]
\centering
\includegraphics[width=0.6\linewidth]{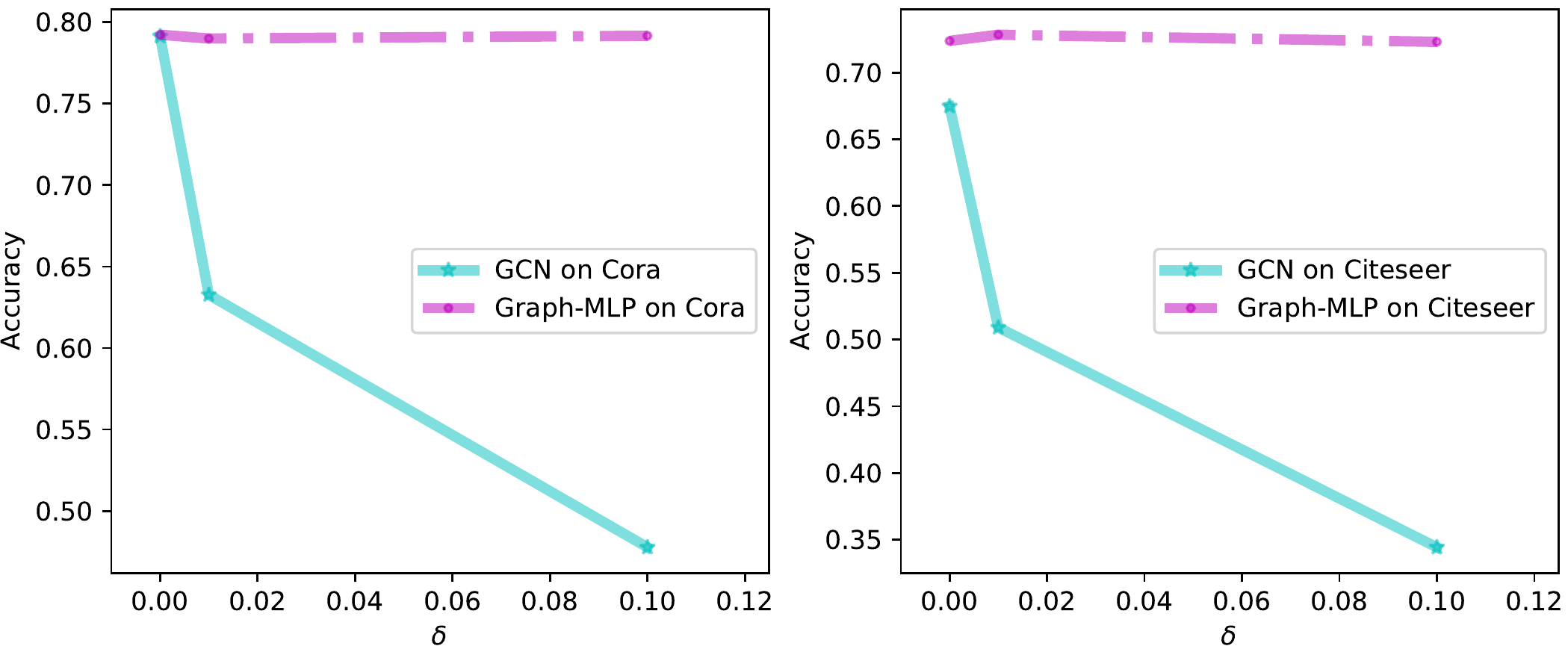}
\caption{Results with corruption on adjacency matrix. For every $\delta$, both GCN and Graph-MLP are trained and tested three times, and averaged accuracy is shown. }
\label{corrupt}
% \vspace{1mm}
\end{figure}

In real world applications, it oftens happens that, testing node features and corrupted adjacency information are provided in inference.  Then, how will the GCN and Graph-MLP behave in the  testing case with corrupted adjacency information? To simulate the corrupted data that may exist in the real word scenarios, we add noises on the adjacency matrix in testing. We define a corruption ratio $\delta$ and the corruption $mask$ to determine positions in the adjacency matrix that need to be corrupted. The corrupted adjacency matrix can be formulated as: 

% Here we define the corruption $mask$ (determine the location of adjacency matrix to be corrupted) as a binary matrix which each its item has p($\delta$) with 0. The corrupted adjacency matrix can be fomulated as:

\begin{equation}
    A_{corr} = A \otimes mask + (1-mask)\otimes \mathbb{N} 
\end{equation}

\begin{equation}
         mask \begin{cases}
= 1,& p=1-\delta \\
= 0,& p=\delta\\
\end{cases}
\end{equation}

% \begin{equation}
%          \mathbb{N} \begin{cases}
% = 1,& p=0.5 \\
% = 0,& p=0.5\\
% \end{cases}
% \end{equation}
where values in $mask \in n \times n$ will be 1/0 under the probability of $1-\delta$/$\delta$. Each item in $\mathbb{N} \in n \times n $ will also be 1/0 under the probability of 0.5/0.5. In total, the formula means that we randomly add or remove edges for a $\delta$\ ratio of the adjacency connections, in order to mimic the noises. In our experiments, we try with $\delta$ to be 0.01 or 0.1 and compare the testing performance of GCN and Graph-MLP. In Figure \ref{corrupt}, we can see that Graph-MLP's performance is free from the influence of corruption in the adjacency matrix. But even a tiny disturbance (setting $\delta$ equals 0.01 so only 1 percent of adjacency information is corrupted) will result in the drastic performance degradation with GCN. Therefore, the Graph-MLP shows great robustness against the corruption in adjacency information.

% Each item in $\mathbb{N} \in n \times n $ distributes uniformly in $U(0,1)$. $\mathbb{N}$ is to simulate the noise level in the chosen corrupted position with $mask_{ij}=0$. Here we test with GCN with the $\widehat{A}_{corr}$ to perform message passing. We compare the GCN with its counterpart Graph-MLP in testing accuracy. In Figure \ref{corrupt}, we can see Graph-MLP's performance are free from the influence of corruption in adjacency matrix, but even tiny disturbance (with $\delta$ equals 0.01, only 1 percent of adjacency information is corrupted) will result in drastic performance degradation on GCN. So, if we use Graph-MLP for inference nodes, we can elude the harmful effect caused by corruption in adjacency information for the robustness our algorithm has.

\section{Conclusion}

In this paper, we propose a novel MLP-based method, Graph-MLP for learning graph node feature distribution. Despite its light structure, to our best knowledge, it is the first deep learning framework to effectively perform graph node classification task without explicit message passing modules. For further supervising our model in term of learning graph node transformation, we propose the NContrast loss to enable learning the structural node distribution without explicit referencing to graph connections \textit{i.e.} adjacency matrix in feedforward. This flexibility in learning contextual knowledge without the message passing modules enables our mode to inference with even corrupted edge information. Our extensive experiments showcase that Graph-MLP is able to deliver comparable and even superior results against the state-of-the-art in node classification tasks with a much simpler and lighter structure.

\bibliographystyle{unsrtnat}
\bibliography{refs}

\begin{thebibliography}{36}
\providecommand{\natexlab}[1]{#1}
\providecommand{\url}[1]{\texttt{#1}}
\expandafter\ifx\csname urlstyle\endcsname\relax
  \providecommand{\doi}[1]{doi: #1}\else
  \providecommand{\doi}{doi: \begingroup \urlstyle{rm}\Url}\fi

\bibitem[Xu et~al.(2017)Xu, Zhu, Choy, and Fei-Fei]{xu2017scene}
Danfei Xu, Yuke Zhu, Christopher~B Choy, and Li~Fei-Fei.
\newblock Scene graph generation by iterative message passing.
\newblock In \emph{Proceedings of the IEEE conference on computer vision and
  pattern recognition}, pages 5410--5419, 2017.

\bibitem[Yang et~al.(2018)Yang, Lu, Lee, Batra, and Parikh]{yang2018graph}
Jianwei Yang, Jiasen Lu, Stefan Lee, Dhruv Batra, and Devi Parikh.
\newblock Graph r-cnn for scene graph generation.
\newblock In \emph{Proceedings of the European conference on computer vision
  (ECCV)}, pages 670--685, 2018.

\bibitem[Zhang et~al.(2018)Zhang, Shi, Xie, Ma, King, and Yeung]{zhang2018gaan}
Jiani Zhang, Xingjian Shi, Junyuan Xie, Hao Ma, Irwin King, and Dit-Yan Yeung.
\newblock Gaan: Gated attention networks for learning on large and
  spatiotemporal graphs.
\newblock \emph{arXiv preprint arXiv:1803.07294}, 2018.

\bibitem[Li et~al.(2017)Li, Yu, Shahabi, and Liu]{li2017diffusion}
Yaguang Li, Rose Yu, Cyrus Shahabi, and Yan Liu.
\newblock Diffusion convolutional recurrent neural network: Data-driven traffic
  forecasting.
\newblock \emph{arXiv preprint arXiv:1707.01926}, 2017.

\bibitem[He et~al.(2020{\natexlab{a}})He, Deng, Wang, Li, Zhang, and
  Wang]{he2020lightgcn}
Xiangnan He, Kuan Deng, Xiang Wang, Yan Li, Yongdong Zhang, and Meng Wang.
\newblock Lightgcn: Simplifying and powering graph convolution network for
  recommendation.
\newblock In \emph{Proceedings of the 43rd International ACM SIGIR Conference
  on Research and Development in Information Retrieval}, pages 639--648,
  2020{\natexlab{a}}.

\bibitem[Kipf and Welling(2016)]{kipf2016semi}
Thomas~N Kipf and Max Welling.
\newblock Semi-supervised classification with graph convolutional networks.
\newblock \emph{arXiv preprint arXiv:1609.02907}, 2016.

\bibitem[Veli{\v{c}}kovi{\'c} et~al.(2017)Veli{\v{c}}kovi{\'c}, Cucurull,
  Casanova, Romero, Lio, and Bengio]{velivckovic2017graph}
Petar Veli{\v{c}}kovi{\'c}, Guillem Cucurull, Arantxa Casanova, Adriana Romero,
  Pietro Lio, and Yoshua Bengio.
\newblock Graph attention networks.
\newblock \emph{arXiv preprint arXiv:1710.10903}, 2017.

\bibitem[Wu et~al.(2019)Wu, Zhang, de~Souza Jr.~au2, Fifty, Yu, and
  Weinberger]{wu2019simplifying}
Felix Wu, Tianyi Zhang, Amauri~Holanda de~Souza Jr.~au2, Christopher Fifty, Tao
  Yu, and Kilian~Q. Weinberger.
\newblock Simplifying graph convolutional networks, 2019.

\bibitem[You et~al.(2020)You, Ying, and Leskovec]{you2020design}
Jiaxuan You, Zhitao Ying, and Jure Leskovec.
\newblock Design space for graph neural networks.
\newblock \emph{Advances in Neural Information Processing Systems}, 33, 2020.

\bibitem[Feng et~al.(2019)Feng, You, Zhang, Ji, and Gao]{feng2019hypergraph}
Yifan Feng, Haoxuan You, Zizhao Zhang, Rongrong Ji, and Yue Gao.
\newblock Hypergraph neural networks.
\newblock In \emph{Proceedings of the AAAI Conference on Artificial
  Intelligence}, volume~33, pages 3558--3565, 2019.

\bibitem[Chami et~al.(2019)Chami, Ying, R{\'e}, and
  Leskovec]{chami2019hyperbolic}
Ines Chami, Rex Ying, Christopher R{\'e}, and Jure Leskovec.
\newblock Hyperbolic graph convolutional neural networks.
\newblock \emph{Advances in neural information processing systems},
  32:\penalty0 4869, 2019.

\bibitem[Liu et~al.(2019)Liu, Nickel, and Kiela]{liu2019hyperbolic}
Qi~Liu, Maximilian Nickel, and Douwe Kiela.
\newblock Hyperbolic graph neural networks, 2019.

\bibitem[Tolstikhin et~al.(2021)Tolstikhin, Houlsby, Kolesnikov, Beyer, Zhai,
  Unterthiner, Yung, Steiner, Keysers, Uszkoreit, Lucic, and
  Dosovitskiy]{tolstikhin2021mlpmixer}
Ilya Tolstikhin, Neil Houlsby, Alexander Kolesnikov, Lucas Beyer, Xiaohua Zhai,
  Thomas Unterthiner, Jessica Yung, Andreas Steiner, Daniel Keysers, Jakob
  Uszkoreit, Mario Lucic, and Alexey Dosovitskiy.
\newblock Mlp-mixer: An all-mlp architecture for vision, 2021.

\bibitem[Melas-Kyriazi(2021)]{melaskyriazi2021need}
Luke Melas-Kyriazi.
\newblock Do you even need attention? a stack of feed-forward layers does
  surprisingly well on imagenet, 2021.

\bibitem[Dosovitskiy et~al.(2020)Dosovitskiy, Beyer, Kolesnikov, Weissenborn,
  Zhai, Unterthiner, Dehghani, Minderer, Heigold, Gelly, Uszkoreit, and
  Houlsby]{dosovitskiy2020image}
Alexey Dosovitskiy, Lucas Beyer, Alexander Kolesnikov, Dirk Weissenborn,
  Xiaohua Zhai, Thomas Unterthiner, Mostafa Dehghani, Matthias Minderer, Georg
  Heigold, Sylvain Gelly, Jakob Uszkoreit, and Neil Houlsby.
\newblock An image is worth 16x16 words: Transformers for image recognition at
  scale, 2020.

\bibitem[Touvron et~al.(2021)Touvron, Bojanowski, Caron, Cord, El-Nouby, Grave,
  Joulin, Synnaeve, Verbeek, and Jégou]{touvron2021resmlp}
Hugo Touvron, Piotr Bojanowski, Mathilde Caron, Matthieu Cord, Alaaeldin
  El-Nouby, Edouard Grave, Armand Joulin, Gabriel Synnaeve, Jakob Verbeek, and
  Hervé Jégou.
\newblock Resmlp: Feedforward networks for image classification with
  data-efficient training, 2021.

\bibitem[Ding et~al.(2021)Ding, Zhang, Han, and Ding]{ding2021repmlp}
Xiaohan Ding, Xiangyu Zhang, Jungong Han, and Guiguang Ding.
\newblock Repmlp: Re-parameterizing convolutions into fully-connected layers
  for image recognition, 2021.

\bibitem[Chen et~al.(2020{\natexlab{a}})Chen, Kornblith, Norouzi, and
  Hinton]{chen2020simple}
Ting Chen, Simon Kornblith, Mohammad Norouzi, and Geoffrey Hinton.
\newblock A simple framework for contrastive learning of visual
  representations.
\newblock In \emph{International conference on machine learning}, pages
  1597--1607. PMLR, 2020{\natexlab{a}}.

\bibitem[Chen et~al.(2020{\natexlab{b}})Chen, Kornblith, Swersky, Norouzi, and
  Hinton]{chen2020big}
Ting Chen, Simon Kornblith, Kevin Swersky, Mohammad Norouzi, and Geoffrey
  Hinton.
\newblock Big self-supervised models are strong semi-supervised learners,
  2020{\natexlab{b}}.

\bibitem[He et~al.(2020{\natexlab{b}})He, Fan, Wu, Xie, and
  Girshick]{he2020momentum}
Kaiming He, Haoqi Fan, Yuxin Wu, Saining Xie, and Ross Girshick.
\newblock Momentum contrast for unsupervised visual representation learning,
  2020{\natexlab{b}}.

\bibitem[Chen et~al.(2020{\natexlab{c}})Chen, Fan, Girshick, and
  He]{chen2020improved}
Xinlei Chen, Haoqi Fan, Ross Girshick, and Kaiming He.
\newblock Improved baselines with momentum contrastive learning,
  2020{\natexlab{c}}.

\bibitem[Grill et~al.(2020)Grill, Strub, Altché, Tallec, Richemond,
  Buchatskaya, Doersch, Pires, Guo, Azar, Piot, Kavukcuoglu, Munos, and
  Valko]{grill2020bootstrap}
Jean-Bastien Grill, Florian Strub, Florent Altché, Corentin Tallec, Pierre~H.
  Richemond, Elena Buchatskaya, Carl Doersch, Bernardo~Avila Pires,
  Zhaohan~Daniel Guo, Mohammad~Gheshlaghi Azar, Bilal Piot, Koray Kavukcuoglu,
  Rémi Munos, and Michal Valko.
\newblock Bootstrap your own latent: A new approach to self-supervised
  learning, 2020.

\bibitem[Khosla et~al.(2021)Khosla, Teterwak, Wang, Sarna, Tian, Isola,
  Maschinot, Liu, and Krishnan]{khosla2021supervised}
Prannay Khosla, Piotr Teterwak, Chen Wang, Aaron Sarna, Yonglong Tian, Phillip
  Isola, Aaron Maschinot, Ce~Liu, and Dilip Krishnan.
\newblock Supervised contrastive learning, 2021.

\bibitem[Gunel et~al.(2020)Gunel, Du, Conneau, and
  Stoyanov]{gunel2020supervised}
Beliz Gunel, Jingfei Du, Alexis Conneau, and Ves Stoyanov.
\newblock Supervised contrastive learning for pre-trained language model
  fine-tuning.
\newblock \emph{arXiv preprint arXiv:2011.01403}, 2020.

\bibitem[Jin et~al.(2020)Jin, Derr, Liu, Wang, Wang, Liu, and
  Tang]{jin2020self}
Wei Jin, Tyler Derr, Haochen Liu, Yiqi Wang, Suhang Wang, Zitao Liu, and
  Jiliang Tang.
\newblock Self-supervised learning on graphs: Deep insights and new direction.
\newblock \emph{arXiv preprint arXiv:2006.10141}, 2020.

\bibitem[Li et~al.(2015)Li, Tarlow, Brockschmidt, and Zemel]{li2015gated}
Yujia Li, Daniel Tarlow, Marc Brockschmidt, and Richard Zemel.
\newblock Gated graph sequence neural networks.
\newblock \emph{arXiv preprint arXiv:1511.05493}, 2015.

\bibitem[Chung et~al.(2014)Chung, Gulcehre, Cho, and
  Bengio]{chung2014empirical}
Junyoung Chung, Caglar Gulcehre, KyungHyun Cho, and Yoshua Bengio.
\newblock Empirical evaluation of gated recurrent neural networks on sequence
  modeling.
\newblock \emph{arXiv preprint arXiv:1412.3555}, 2014.

\bibitem[Hendrycks and Gimpel(2016)]{hendrycks2016gaussian}
Dan Hendrycks and Kevin Gimpel.
\newblock Gaussian error linear units (gelus).
\newblock \emph{arXiv preprint arXiv:1606.08415}, 2016.

\bibitem[Ba et~al.(2016)Ba, Kiros, and Hinton]{ba2016layer}
Jimmy~Lei Ba, Jamie~Ryan Kiros, and Geoffrey~E Hinton.
\newblock Layer normalization.
\newblock \emph{arXiv preprint arXiv:1607.06450}, 2016.

\bibitem[Srivastava et~al.(2014)Srivastava, Hinton, Krizhevsky, Sutskever, and
  Salakhutdinov]{srivastava2014dropout}
Nitish Srivastava, Geoffrey Hinton, Alex Krizhevsky, Ilya Sutskever, and Ruslan
  Salakhutdinov.
\newblock Dropout: a simple way to prevent neural networks from overfitting.
\newblock \emph{The journal of machine learning research}, 15\penalty0
  (1):\penalty0 1929--1958, 2014.

\bibitem[Hamilton et~al.(2017)Hamilton, Ying, and
  Leskovec]{hamilton2017inductive}
William~L Hamilton, Rex Ying, and Jure Leskovec.
\newblock Inductive representation learning on large graphs.
\newblock \emph{arXiv preprint arXiv:1706.02216}, 2017.

\bibitem[Kingma and Ba(2014)]{kingma2014adam}
Diederik~P Kingma and Jimmy Ba.
\newblock Adam: A method for stochastic optimization.
\newblock \emph{arXiv preprint arXiv:1412.6980}, 2014.

\bibitem[Liao et~al.(2019)Liao, Zhao, Urtasun, and Zemel]{liao2019lanczosnet}
Renjie Liao, Zhizhen Zhao, Raquel Urtasun, and Richard~S. Zemel.
\newblock Lanczosnet: Multi-scale deep graph convolutional networks, 2019.

\bibitem[Perozzi et~al.(2014)Perozzi, Al-Rfou, and Skiena]{perozzi2014deepwalk}
Bryan Perozzi, Rami Al-Rfou, and Steven Skiena.
\newblock Deepwalk: Online learning of social representations.
\newblock In \emph{Proceedings of the 20th ACM SIGKDD international conference
  on Knowledge discovery and data mining}, pages 701--710, 2014.

\bibitem[Velickovic et~al.(2019)Velickovic, Fedus, Hamilton, Li{\`o}, Bengio,
  and Hjelm]{velickovic2019deep}
Petar Velickovic, William Fedus, William~L Hamilton, Pietro Li{\`o}, Yoshua
  Bengio, and R~Devon Hjelm.
\newblock Deep graph infomax.
\newblock In \emph{ICLR (Poster)}, 2019.

\bibitem[Van~der Maaten and Hinton(2008)]{van2008visualizing}
Laurens Van~der Maaten and Geoffrey Hinton.
\newblock Visualizing data using t-sne.
\newblock \emph{Journal of machine learning research}, 9\penalty0 (11), 2008.

\end{thebibliography}

\end{document}